\documentclass{article} % For LaTeX2e
\usepackage{iclr2024_conference,times}
\usepackage{graphicx}
\usepackage{amsmath}
\usepackage{amssymb}
\usepackage{booktabs}
\usepackage{booktabs}       % professional-quality tables
\usepackage{amsfonts}       % blackboard math symbols
\usepackage{nicefrac}       % compact symbols for 1/2, etc.
\usepackage{microtype}      % microtypography
\usepackage{xcolor}         % colors
\usepackage{wrapfig,lipsum,booktabs}

\usepackage[colorlinks,
            linkcolor=black,       %%修改此处为你想要的颜色
            anchorcolor=black,  %%修改此处为你想要的颜色
            citecolor=cyan,        %%修改此处为你想要的颜色，例如修改blue为red
            ]{hyperref}
\usepackage{amsfonts,amssymb}
\usepackage{wrapfig}
\usepackage{url}
\usepackage{graphicx}
\usepackage{algorithmic}
\usepackage{algorithm}
\usepackage{color}

\usepackage{graphicx}
\usepackage{float}
\usepackage{subfigure}
\usepackage{tcolorbox}
\usepackage{times}
\usepackage{helvet}
\usepackage{courier}
\usepackage{algorithmic}
\usepackage{algorithm}
\usepackage{color}

\newcommand{\myparagraph}[1]{\vspace{0.0em}\noindent\vspace{0.0em}\textbf{#1}}
\usepackage{amsmath,amsfonts,bm}

\def\eqref#1{equation~\ref{#1}}
\def\1{\bm{1}}

\DeclareMathAlphabet{\mathsfit}{\encodingdefault}{\sfdefault}{m}{sl}
\SetMathAlphabet{\mathsfit}{bold}{\encodingdefault}{\sfdefault}{bx}{n}

\usepackage{diagbox}
\usepackage{hyperref}
\usepackage{marvosym}
\usepackage{url}
\usepackage{makecell}
\usepackage{adjustbox}
\usepackage{url}
\title{CSGO: Content-Style Composition in Text-to-Image Generation}

\author{Peng Xing$^{*12}$, Haofan Wang$^{*2}$, Yanpeng Sun$^{1}$, Qixun Wang$^{2}$, \\
\textbf{Xu Bai$^{2,3}$, Hao Ai$^{2}$, Renyuan Huang$^{2}$, Zechao Li\textsuperscript{\Letter}$^{1}$}\\
Nanjing University of Science and Technology, InstantX Team, Xiaohongshu Inc\\
\texttt{\{xingp\_ng, zechao.li\}@njust.edu.cn}, \texttt{haofanwang.ai@gmail.com}\\
}

\iclrfinalcopy % Uncomment for camera-ready version, but NOT for submission.
\begin{document}

\maketitle

\begin{figure*}[h]
    \centering\setlength{\abovecaptionskip}{2pt}
    \includegraphics[width=.99\linewidth]{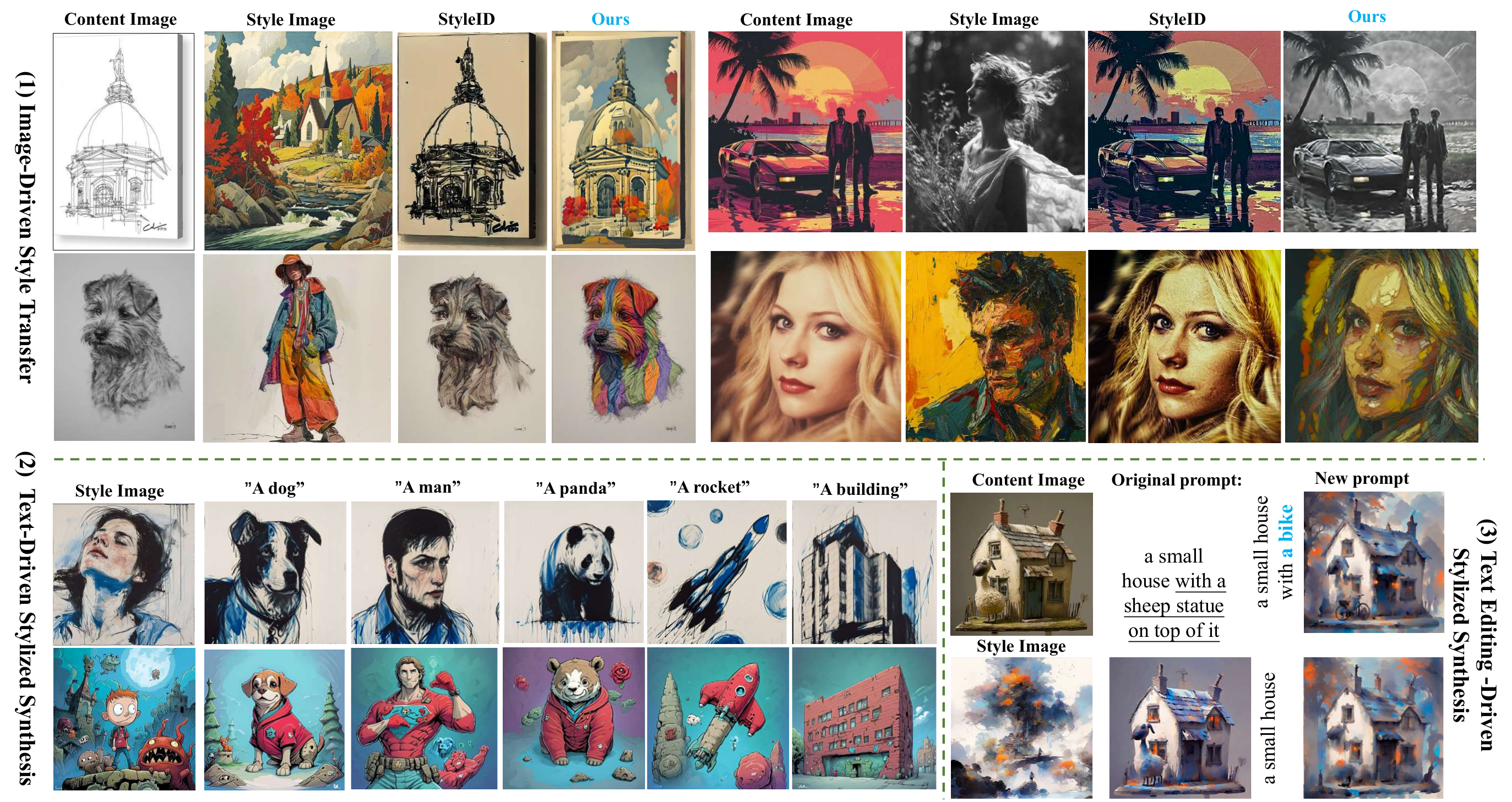}
    \caption{
    (1) Comparison of the style transfer results of the proposed method with the recent state-of-the-art method StyleID~\cite{chung2024style}.
    (2) Our CSGO achieves high-quality text-driven stylized synthesis. (3) Our CSGO achieves high-quality text editing-driven stylized synthesis.}
    \label{fig1}
\end{figure*}

\begin{abstract}
The diffusion model has shown exceptional capabilities in controlled image generation, which has further fueled interest in image style transfer.
Existing works mainly focus on training free-based methods (e.g., image inversion) due to the scarcity of specific data.
In this study, we present a data construction pipeline for content-style-stylized image triplets that generates and automatically cleanses stylized data triplets. 
Based on this pipeline, we construct a dataset IMAGStyle, the first large-scale style transfer dataset containing 210k image triplets, available for the community to explore and research.
Equipped with IMAGStyle, we propose CSGO, a style transfer model based on end-to-end training, which explicitly decouples content and style features employing independent feature injection. 
The unified CSGO implements image-driven style transfer, text-driven stylized synthesis, and text editing-driven stylized synthesis.
Extensive experiments demonstrate the effectiveness of our approach in enhancing style control capabilities in image generation.
Additional visualization and access to the source code can be located on the project page: \url{https://csgo-gen.github.io/}.
\end{abstract}

% \begin{figure*}[h]
%     \centering\setlength{\abovecaptionskip}{2pt}
%     \includegraphics[width=.99\linewidth]{pictures/image1-3.pdf}
%     \caption{
%     %
%     (1) Comparison of the style transfer results of the proposed method with the recent state-of-the-art method StyleID~\cite{chung2024style}.
%     %
%     (2) The proposed CSGO achieves high-quality text-driven stylized synthesis. (3) The proposed CSGO achieves high-quality text-driven image editing.}
%     \label{fig1}
% \end{figure*}
% \section{Introduction}

%
% Early, researchers attempted to solve the above challenges by utilizing paired~\cite{isola2017image} or unpaired data~\cite{zhu2017unpaired} through convolution neural networks~\cite{gatys2016image,park2019arbitrary,gatys2017controlling}, GAN, transformer structures~\cite{li2019learning}. 

\section{Introduction}
Recent advancements in diffusion models have significantly improved the field of text-to-image generation~\cite{song2020denoising,ho2020denoising}. 
Models such as SD~\cite{DBLP:conf/cvpr/RombachBLEO22} excel at creating visually appealing images based on textual prompts, playing a crucial role in personalized content creation~\cite{ruiz2023dreambooth,xu2024imagereward}.
%
% Models such as SD~\cite{DBLP:conf/cvpr/RombachBLEO22}, SDXL~\cite{DBLP:journals/corr/abs-2307-01952} and DALL-E2~\cite{ramesh2021zero}, excel at creating visually appealing images based on textual prompts, playing a crucial role in personalized content creation~\cite{ruiz2023dreambooth,xu2024imagereward}, image editing~\cite{mokady2023null,hertz2022prompt} and style control~\cite{wang2024instantstyle,rout2024rb}. 
%
Despite numerous studies on general controllability, image style transfer remains particularly challenging.

Image style transfer aims to generate a plausible target image by combining the content of one image with the style of another, ensuring that the target image maintains the original content's semantics while adopting the desired style~\cite{jing2019neural,deng2020arbitrary}.
This process requires fine-grained control over content and style, involving abstract concepts like texture, color, composition, and visual quality, making it a complex and nuanced challenge~\cite{chung2024style}.

A significant challenge in style transfer is the lack of a large-scale stylized dataset, which makes it impossible to train models end-to-end and results in suboptimal style transfer quality for non-end-to-end methods.
%
% Existing methods are typically training-free structures that utilize DDIM inversion~\cite{song2020denoising} or carefully tuned feature injection layers of pre-trained IP-Adapter~\cite{ye2023ip} to achieve style transfer.
%
Existing methods typically rely on training-free structures, such as DDIM inversion~\cite{song2020denoising} or carefully tuned feature injection layers of pre-trained IP-Adapter~\cite{ye2023ip}. 
Methods like Plug-and-Play~\cite{tumanyan2023plug}, VCT~\cite{cheng2023general}, and the state-of-the-art StyleID~\cite{chung2024style} employ content image inversion and sometimes style image inversion to extract and inject image features into specifically designed layers.
However, inverting content and style images significantly increases inference time, and DDIM inversion can lose critical information~\cite{mokady2023null}, leading to failures, as shown in Figure~\ref{fig1}.
InstantStyle~\cite{wang2024instantstyle} employs the pre-trained IP-Adapter. However, it struggles with accurate content control.
Another class of methods relies on a small amount of data to train LoRA and implicitly decouple content and style LoRAs, such as ZipLoRA~\cite{shah2023ziplora} and B-LoRA~\cite{frenkel2024implicit}, which combine style and content LoRAs to achieve content retention and Style transfer. 
However, each image requires fine-tuning, and implicit decoupling reduces stability.

To overcome the above challenges, we start by constructing a style transfer-specific dataset and then design a simple yet effective framework to validate the beneficial effects of this large-scale dataset on style transfer. 
Initially, we propose a dataset construction pipeline for Content-Style-Stylized Image Triplets (CSSIT), incorporating both a data generation method and an automated cleaning process.
Using this pipeline, we construct a large-scale stylized dataset, \textbf{IMAGStyle}, comprising 210K content-style-stylized image triplets.
Next, we introduce an end-to-end trained style transfer framework, \textbf{CSGO}. Unlike previous implicit extractions, it explicitly uses independent content and style feature injection modules to achieve high-quality image style transformations.
The framework simultaneously accepts style and content images as inputs and efficiently fuses content and style features using well-designed feature injection blocks.
Benefiting from the decoupled training framework, once trained, CSGO realizes any form of arbitrary style transfer without fine-tuning at the inference stage, including sketch or nature image-driven style transfer, text-driven, text editing-driven stylized synthesis.
Finally, we introduce a Content Alignment Score (CAS) to evaluate the quality of style transfer, effectively measuring the degree of content loss post-transfer.
Extensive qualitative and quantitative studies validate that our proposed method achieves advanced zero-shot style transfer.
%三点可写可不写  数据构造方式 数据集
% 框架
%指标 和结果
%#########################################################%

\section{Related Work}
\myparagraph{Text-to-Image Model.}
In recent years, diffusion models have garnered significant attention in the text-to-image generation community due to their powerful generative capabilities demonstrated by early works~\cite{dhariwal2021diffusion,ramesh2022hierarchical}.
Owing to large-scale training~\cite{schuhmann2022laion}, improved architectures~\cite{radford2021learning,DBLP:conf/iccv/PeeblesX23}, and latent space diffusion mechanisms, models like Stable Diffusion have achieved notable success in text-to-image generation.~\cite{ramesh2022hierarchical}.
The focus on controllability in text-to-image models has grown in response to practical demands. Popular models such as ControlNet~\cite{zhang2023adding}, T2Iadapter~\cite{mou2024t2i}, and IP-Adapter~\cite{ye2023ip} introduce additional image conditions to enhance controllability. These models use sophisticated feature extraction methods and integrate these features into well-designed modules to achieve layout control.
In this paper, we present a style transfer framework, CSGO, based on an image-conditional generation model that can perform zero-shot style transfer.

\myparagraph{Style Transfer.}
Style transfer has garnered significant attention and research due to its practical applications in art creation~\cite{gatys2016image}.
Early methods, both optimization-based~\cite{gatys2016image} and inference-based~\cite{chen2017stylebank,dumoulin2016learned}, are limited by speed constraints and the diversity of style transfer. 
% Early optimization method~\cite{gatys2016image} and inference methods~\cite{chen2017stylebank,dumoulin2016learned} are limited by speed constraints and the diversity of style transfer.
%
The AdaIN approach~\cite{huang2017arbitrary}, which separates content and style features from deep features, has become a representative method for style transfer, inspiring a series of techniques using statistical mean and variance~\cite{chen2021artistic,hertz2024style}.
Additionally, transformer-based methods such as StyleFormer~\cite{wu2021styleformer} and StyTR$^2$~\cite{deng2022stytr2} improve content bias. 
However, these methods primarily focus on color or stroke transfer and face limitations in arbitrary style transfer.

Currently, inversion-based Style Transfer (InST)~\cite{zhang2023inversion} is proposed to obtain inversion latent of style image and manipulate attention maps to edit generated Images.
However, DDIM (Denoising Diffusion Implicit Models) inversion results in content loss and increased inference time~\cite{song2020denoising}. 
Hertz~\textit{et al.} explore self-attention layers using key and value matrices for style transfer~\cite{hertz2024style}. 
DEADiff~\cite{qi2024deadiff} and StyleShot~\cite{gao2024styleshot} are trained through a two-stage style control method. However, it is easy to lose detailed information within the control through sparse lines.
InstantStyle~\cite{wang2024instantstyle,wang2024instantstyleplus} to achieve high-quality style control through pre-trained prompt adapter~\cite{ye2023ip} and carefully designed injection layers.
However, these methods struggle with achieving high-precision style transfer and face limitations related to content preservation. 
Some fine-tuning approaches, such as LoRA~\cite{hu2021lora}, DB-LoRA~\cite{ryulow}, Zip-LoRA~\cite{shah2023ziplora}, and B-LoRA~\cite{frenkel2024implicit}, enable higher-quality style-controlled generation but require fine-tuning for different styles and face challenges in achieving style transfer. 
Our proposed method introduces a novel style transfer dataset and develops the CSGO framework, achieving high-quality arbitrary image style transfer without the need for fine-tuning.

\section{Data Pipeline}

In this section, we first introduce the proposed pipeline for constructing content-style-stylized image triplets. Then,  we describe the constructed IMAGStyle dataset in detail.

\begin{wrapfigure}{r}{0.5\textwidth}
% \begin{figure}
% \begin{minipage}
\vspace{-0.6cm}
    \centering\setlength{\abovecaptionskip}{0pt}
    \includegraphics[width=.49\textwidth]{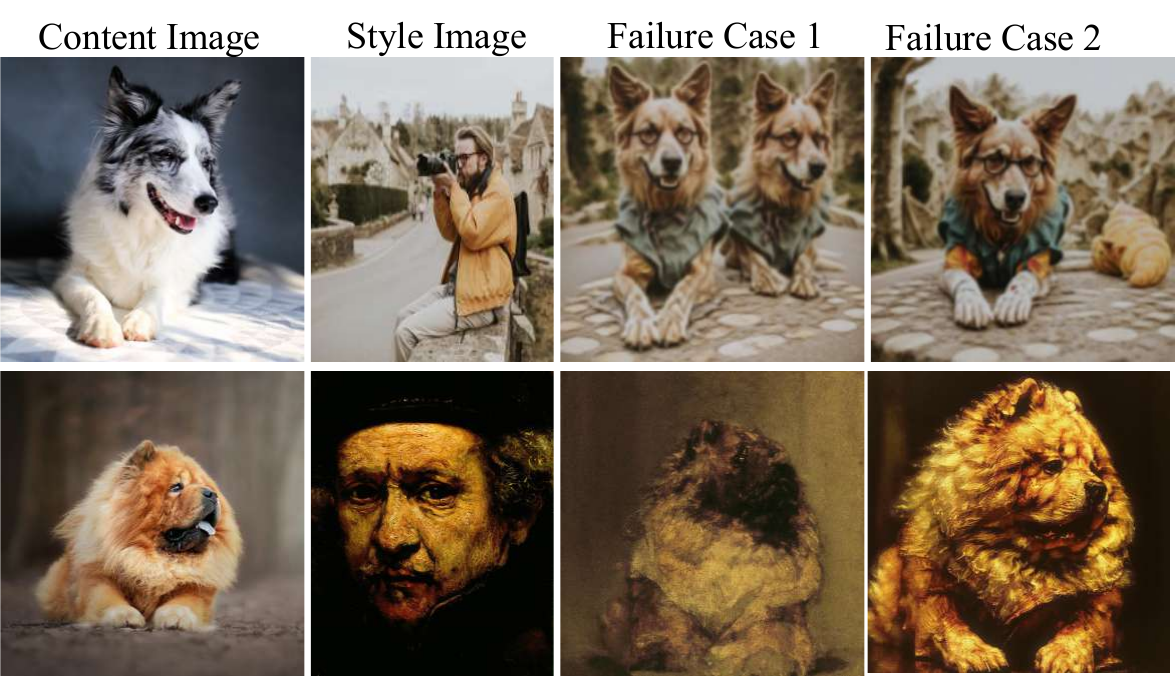}
    % \vspace{-6pt}
    \caption{Failure cases in step (1), which fail to maintain the spatial structure of the content image.}
    \label{fig:2}
    \vspace{-0.8cm}
    % \end{minipage}
\end{wrapfigure}

\subsection{Pipeline for Constructing Content-Style-Stylized Image Triplets}\label{sec3.1}

% One of the challenges of style transfer is the lack of data pairs of content-style- stylized images. 
%
The lack of a large-scale open-source dataset of content-style-stylized image pairs (CSSIT) in the community seriously hinders the research on style transfer. 
In this work, we propose a data construction pipeline that automatically constructs and cleans to obtain high-quality content-style-stylized image triplets, given only arbitrary content images and style images. 
The pipeline contains two steps: (1) stylized image generation and (2) stylized image cleaning.

\begin{wrapfigure}{r}{0.5\textwidth}
    \vspace{-3pt}
    \begin{minipage}{0.5\textwidth}
        \begin{algorithm}[H]
            \scriptsize
            \caption{Pipeline of Constructing CSSIT}
            \label{alg}
            \renewcommand{\algorithmicrequire}{\textbf{Input:}}
            \renewcommand{\algorithmicensure}{\textbf{Output:}}
            \begin{algorithmic}[1]
                \REQUIRE content images $Set_{content}$, style images $Set_{style}$ % input
                \ENSURE Content-style-stylized image triplets $Set$   % output
                \FOR{each $C \in Set_{content}$}
                    \STATE $C_{LoRA} \leftarrow$ Train LoRA for $C$
                    \STATE $C_{LoRA}^{content} \leftarrow$ Separate content LoRA in $C_{LoRA}$ 
                    \FOR{each $S \in Set_{style}$}
                        \STATE $S_{LoRA} \leftarrow$ Train LoRA for $S$   
                        \STATE $S_{LoRA}^{style} \leftarrow$ Separate style LoRA in $S_{LoRA}$
                        \STATE $CS_{LoRA} \leftarrow $ Combine $C_{LoRA}^{content}$ and $S_{LoRA}^{style}$
                        \STATE $T = \{T_1,T_2,...,T_n\} \leftarrow $ Generate $n$ images by SDXL and $CS_{LoRA}$
                        \STATE $CAS_1,CAS_2,...,CAS_n \leftarrow $ Compute CAS for each generated image based on Equ.(~\ref{equ1})
                        \STATE $i \leftarrow$ Obtain the index of the minimum value of all CAS
                        \STATE $Set$.append([$C,S,T_i$])
                    \ENDFOR
                \ENDFOR
                
                \RETURN $Set$ % \textcolor{blue}{\COMMENT{Each item in $Set$ is a triplet of content-style-stylized images.}}
            \end{algorithmic}
        \end{algorithm}
    \end{minipage}
    \vspace{-10pt}
\end{wrapfigure}

\myparagraph{Stylized image generation. }
% Given an arbitrary content image $C$, and an arbitrary image $S$, our purpose is to obtain a single stylized image $T$ whose content is aligned with $C$, and whose style is similar to $S$.
Given an arbitrary content image $C$ and an arbitrary style image $S$, the goal is to generate a stylized image $T$ that preserves the content of $C$ while adopting the style of $S$.
We are inspired by B-LoRA~\cite{frenkel2024implicit}, which finds that content LoRA and style LoRA can be implicitly separated by SD-trained LoRA, preserving the original image's content and style information, respectively.
%
 % We follow the recent work B-LoRA~\cite{frenkel2024implicit} and train B-LoRA for $C$ and $S$ using a single image $C$ and image $S$, respectively. 
 %
 Therefore, we first train a large number of LoRAs with lots of content and style imges.
 To ensure that the content of the generated image $T$ is aligned to $C$ as much as possible, the loRA for $C$ is trained using only one content image $C$.
 Then, Each trained loRA is decomposed into a content LoRA and a style LoRA through implicit separate mentioned by work~\cite{frenkel2024implicit}. Finally, the content LoRA of image $C$ is combined with the style LoRA of $S$ to generate the target images $T = \{T_1,T_2,...,T_n\}$ using the base model.
 However, the implicit separate approach is unstable, resulting in the content and style LoRA not reliably retaining content or style information.
 This manifests itself in the form of the generated image $T_i$, which does not always agree with the content of $C$, as shown in Figure~\ref{fig:2}. 
 Therefore, it is necessary to filter $T$, sampling the most reasonable $T_i$ as the target image.

 \myparagraph{Stylized image cleaning. } 
 Slow methods of cleaning data with human involvement are unacceptable for the construction of large-scale stylized data triplets. 
 %
 % Therefore, there is an urgent need to develop an automated cleaning method to obtain high-quality style transfer results.
 %
To this end, we develop an automatic cleaning method to obtain the ideal and high-quality stylized image $T$ efficiently. 
 First, we propose a content alignment score (CAS) that effectively measures the content alignment of the generated image with the content image.
 It is defined as the feature distance between the content semantic features (without style information) of the generated image and the original content image. It is represented as follows:
\begin{equation}
    CAS_i=\left \|Ada(\phi(C))-Ada(\phi(T_i))  \right \| ^2,\label{equ1}
\end{equation}
where $CAS_i$ denotes the content alignment score of generated image $T_i$, $\phi(\cdot)$ denotes image encoder. We compare the mainstream feature extractors and the closest to human filtering results is DINO-V2~\cite{li2023blip}. $Ada(F)$ represents a function of feature $F$ to remove style information. 
We follow AdaIN~\cite{huang2017arbitrary} to express style information by mean and variance. It is represented as follows:
 \begin{equation}
     Ada(F) = \frac{F-\mu (F)}{\rho (F)},
 \end{equation}
where $\mu (F)$ and $\rho(F)$ represent the mean and variance of feature $F$. % 
Obviously, a smaller CAS indicates that the generated image is closer to the content of the original image. In Algorithm~\ref{alg}, we provide a pseudo-code of our pipeline.

% \begin{algorithm}[!h]
%     \scriptsize
%     \caption{Pipeline of Constructing CSSIP}
%     \label{alg}
%     \renewcommand{\algorithmicrequire}{\textbf{Input:}}
%     \renewcommand{\algorithmicensure}{\textbf{Output:}}
%     \begin{algorithmic}[1]
%         \REQUIRE content images $Set_{content}$, style images $Set_{style}$ %%input
%         \ENSURE Content-style-stylized image pairs $Set$   %%output
%         \FOR{each $C \in Set_{content}$}
%             \STATE $C_{B-LoRA} \leftarrow$ Train B-LoRA for $C$
%             \STATE $C_{B-LoRA}^{content} \leftarrow$ Separate content LoRA in $C_{B-LoRA}$ 
%             \FOR{each $S \in Set_{style}$}
%                 \STATE $S_{B-LoRA} \leftarrow$ Train B-LoRA for $S$   
%                 \STATE $S_{B-LoRA}^{style} \leftarrow$ Separate style LoRA in $S_{B-LoRA}$
%                 \STATE $CS_{B-LoRA} \leftarrow $ Combine $C_{B-LoRA}^{content}$ and $S_{B-LoRA}^{style}$
%                 \STATE
%                 $G = \{C_S^1,C_S^2,...,C_S^n\}\leftarrow $ Gen(SDXL,$CS_{B-LoRA}$) \textcolor{blue}{\COMMENT{Generate $n$ images based on the diffusion model and $CS_{B-LoRA}$}}
%                 \STATE $CAS_1,CAS_2,...,CAS_n\leftarrow $ Compute CAS for each generated image based Equ. (~\ref{equ1})
%                 \STATE $i \leftarrow$ Obtain the index of minimum value of all CAS
%                 \STATE $Set.append([C,S,C_S^i])$
%             \ENDFOR
%         \ENDFOR
        
%         \RETURN $Set$ \textcolor{blue}{\COMMENT{Each item in $Set$ is pair of content-style-stylized images.}}
%     \end{algorithmic}
% \end{algorithm}

\subsection{IMAGStyle Dataset Details}

\begin{wrapfigure}{r}{0.35\textwidth}
\vspace{-0.35cm}
    \centering\setlength{\abovecaptionskip}{0pt}
    \includegraphics[width=.34\textwidth]{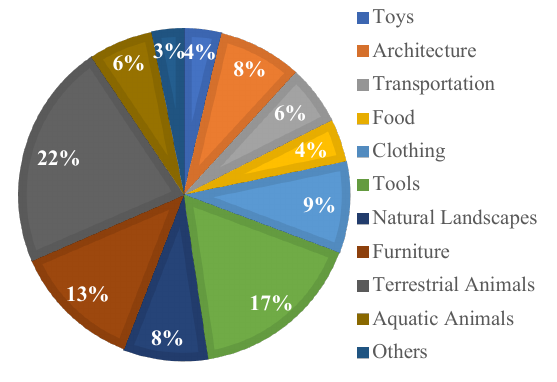}
    % \vspace{-6pt}
    \caption{Distribution of content images.}
    \label{fig:31}
    \vspace{-0.7cm}
    % \end{minipage}
\end{wrapfigure}

\myparagraph{Content Images.} To ensure that the content images have clear semantic information and facilitate separating after training, we employ the saliency detection datasets, MSRA10K~\cite{ChengPAMI,13iccv/Cheng_Saliency} and MSRA-B~\cite{WangDRFI2017}, as the content images. In addition, for sketch stylized, we sample 1000 sketch images from ImageNet-Sketch~\cite{wang2019learning} as content images. The category distribution of content images is shown in Figure~\ref{fig:31}.
We use BLIP~\cite{li2023blip} to generate a caption for each content image.
A total of 11,000 content images are trained and used as content LoRA. 

\myparagraph{Style Images.} 
To ensure the richness of the style diversity, we sample 5000 images of different painting styles (history painting, portrait, genre painting, landscape, and still life) from the Wikiart dataset~\cite{saleh2016large}. In addition, we generated 5000 images using Midjourney covering diverse styles, including Classical, Modern, Romantic, Realistic, Surreal, Abstract, Futuristic, Bright, Dark styles etc. A total of 10,000 style images are used to train style LoRA.

\myparagraph{Dataset.} Based on the pipeline described in Section~\ref{sec3.1}, we construct a style transfer dataset, \textbf{IMAGStyle}, which contains 210K content-style-stylized image triplets as training dataset. Furthermore, we collect 248 content images from the web containing images of real scenes, sketched scenes, faces, and style scenes, as well as 206 style images of different scenes as testing dataset. For testing, each content image is transferred to 206 styles. This dataset will be used for community research on style transfer and stylized synthesis.

\begin{figure*}[t]
    \centering\setlength{\abovecaptionskip}{2pt}
    \includegraphics[width=.99\linewidth]{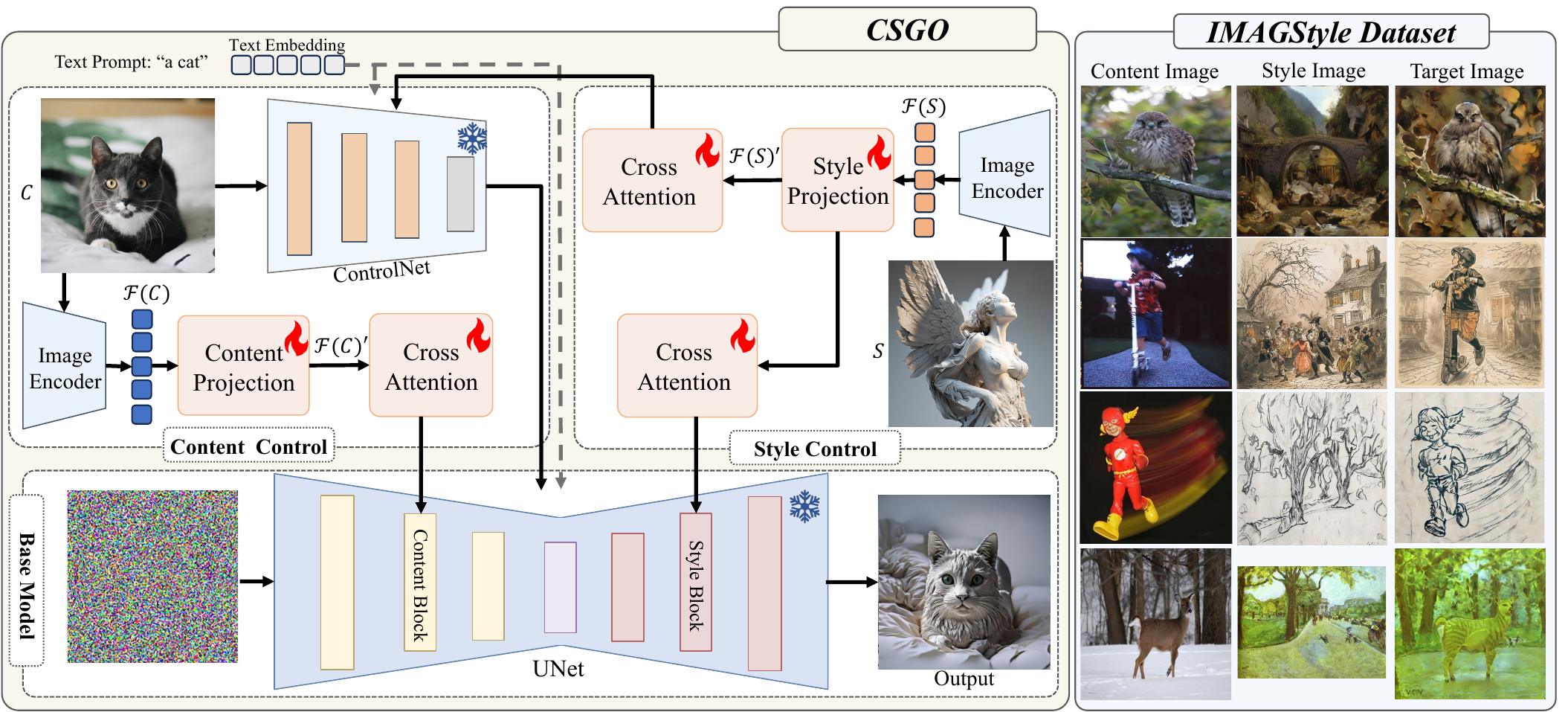}
    \caption{(a) Left: Overview of the proposed end-to-end style transfer framework CSGO.  (b) Right: Samples from our IMAGStyle dataset.
    %
    % First, for content control, we use IPadapter injection and ControlNet injection. ControlNet extracts the content features of content image C and injects them into the up-sampling blocks of the base model. The content IPadapter extracts the semantic information of C and injects it into the down-sampling blocks of the base model. Second, for style control, we use IPadater to extract the style features of the style image S to inject into the up-sampling blocks of ControlNet model and base model respectively. The decoupled block injection approach avoids the entanglement of style and content.
    %
    % Finally, in the inference stage, the styled results can be generated efficiently without fine-tuning.
    }
    \label{fig2}
    \vspace{-0.5cm}
\end{figure*}

%#########################################################%

\section{Approach}

\subsection{CSGO framework}

% \begin{figure}[t]
% \centering  %图片全局居中
% \subfigure[image 1]{
% \label{Fig.sub.1}
% \includegraphics[width=.3\linewidth]{pictures/image3.pdf}}
% \subfigure[image 2]{
% \label{Fig.sub.2}
% \includegraphics[width=.65\linewidth]{pictures/image2.pdf}}
% \caption{Insert two pictures side by side}
% \label{1}
% \end{figure}

The proposed style transfer model, CSGO, shown in Figure~\ref{fig2}, aims to achieve arbitrary stylization of any image without fine-tuning, including sketch and natural image-driven style transfer, text-driven stylized synthesis, and text editing-driven stylized synthesis. 
Benefiting from the proposed IMAGStyle dataset, the proposed CSGO supports an end-to-end style transfer training paradigm. 
To ensure effective style transfer and accurate content preservation, we carefully design the content and style control modules.
In addition, to reduce the risk that the content image leaks style information or the style image leaks content, the content control and style control modules are explicitly decoupled, and the corresponding features are extracted independently.
To be more specific, we categorize our CSGO into two main components and describe them in detail.

% \subsection{Model Architecture}
\myparagraph{Content Control. } The purpose of content control is to ensure that the stylized image retains the semantics, layout, and other features of the content image.
To this end, we carefully designed two ways of content control. 
First, we implement content control through pre-trained ControlNet~\cite{zhang2023adding}, whose input is the content image and the corresponding caption.
We leverage the capabilities of the specific content-controllable model(Tile ControlNet) to reduce the data requirements and computational costs of training content retention from scratch
Following the ControlNet, the output of ControlNet is directly injected into the up-sampling blocks of the base model (pre-trained UNet in SD) to obtain fusion output  $D_i^\prime = D_i+\delta_c \times C_i$, $D_i$ denotes the output of $i$-th block in the base model, $C_i$ denotes the output of $i$-th block in ControlNet, $\delta_c$ represents the fusion weight.
In addition, to achieve content control in the down-sampling blocks of the base model, we utilize an additional learnable cross-attention layer to inject content features into down blocks.
Specifically, we use pre-trained CLIP image encoder~\cite{radford2021learning} and a learnable projection layer to extract the semantic feature $\mathcal{F}(C)^\prime$ of the content image. Then, we utilize an additional cross-attention layer to inject the extracted content features into the down-sampling blocks of the base model, \textit{i.e.}, $D_C^\prime = D+\lambda_c \times D_C$, $D$ denotes the output of in the base model, $D_C$ denotes the output of content IP-Adapter, $\lambda_c$ represents the fusion weight~\cite{ye2023ip}. 
These two content control strategies ensure small content loss during the style transfer.

\myparagraph{Style Control. }
% In style transfer, style control is subject to two conditions, precise transfer of styles and no adjustment to the original semantics.
%
To ensure that the proposed CSGO has strong style control capability, we also design two simple yet effective style control methods.
Generally, we feed the style images into a pre-trained image encoder to extract the original embedding $\mathcal{F}(S) \in \mathbb{R}^{o \times d}$ and map them to the new embedding $\mathcal{F}(S)^\prime \in \mathbb{R}^{t \times d}$ through the Style Projection layer. Here, $o$ and $t$ represent the token number of original and new embeddings, $d$ denotes the dimension of $\mathcal{F}(S)$.
For style projection, we employ the Perceiver Resampler structure~\cite{alayrac2022flamingo} to obtain more detailed style features. 
Then, we utilize an additional cross-attention layer to inject the new embedding into the up-sampling blocks of the base model.
Furthermore, we note that relying only on the injection of the up-sampling blocks of the base model weakens the style control since ControlNet injections in the content control may leak style information of the content image $C$.
For this reason, we propose to use an independent cross attention module to simultaneously inject style features into, and the fusion weight is $\lambda_s$, as shown in Figure~\ref{fig2}.
The insight of this is to pre-adjust the style of the content image using style features making the output of the Controlnet model retain the content while containing the desired style features.

In summary, the proposed CSGO framework explicitly learns separate feature processing modules that inject style and content features into different locations of the base model, respectively. Despite its simplicity, CSGO achieves state-of-the-art style transfer results.

\begin{figure}[ht]
    \centering\setlength{\abovecaptionskip}{2pt}
    \includegraphics[width=.99\linewidth]{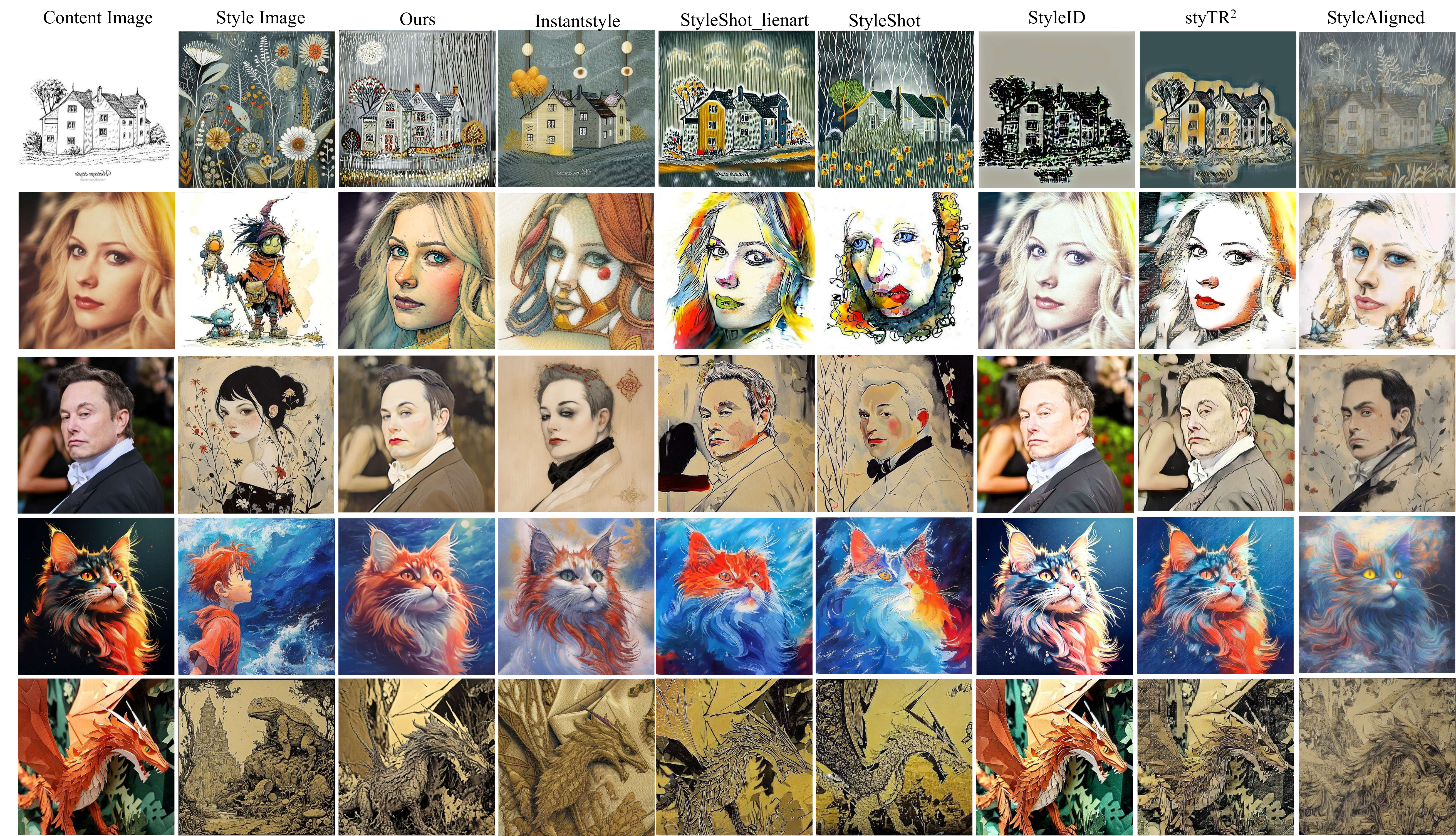}
    \caption{Comparison of image-driven style transfer results. Zoomed in for the best viewing.}
    \label{fig:3}
    \vspace{-0.3cm}
\end{figure}

% \myparagraph{Base Models.}

\subsection{Model Training and Inference. }
\myparagraph{Training.} Based on the proposed dataset, IMAGStyle, our CSGO is the first implementation of end-to-end style transfer training. Given a content image $C$, a caption $P$ of the content image, a style image $S$, and a target image $T$, we train a style transfer network based on a pre-trained diffusion model. Our training objective is to model the relationship between the styled image $T$ and Gaussian noise under content and style image conditions, which is represented as follows:
\begin{equation}
    \mathcal{L}=\mathbb{E}_{z_{0}, t, P, C, S, \epsilon \sim \mathcal{N}(0,1)}\left[\left\|\epsilon-\epsilon_{\theta}\left(z_{t}, t, C,S, P\right)\right\|^{2}\right],
\end{equation}
where $\varepsilon$ denotes the random sampled Gaussian noise, $\varepsilon_\theta$ denotes the trainable parameters of CSGO, $t$ represents the timestep.  Note that the latent latent $z_t$ is constructed with a style image $T$ during training, $z_t=\sqrt{\bar{\alpha}_{t}} \psi (T)+\sqrt{1-\bar{\alpha}_{t}} \varepsilon$, where $\psi(\cdot)$ mapping the original input to the latent space function, $\bar{\alpha_t}$ is consistent with  diffusion models~\cite{song2020denoising,ho2020denoising}. 
We randomly drop content image and style image conditions in the training phase to enable classifier-free guidance in the inference stage. 

\myparagraph{Inference. }
During the inference phase, we employ classifier-free guidance.  The output of timestep $t$ is indicated as follows:
\begin{equation}
\hat{{\epsilon}}_\theta({z}_t,t,{C},{S},P)=w{\epsilon}_\theta({z}_t,t,{C},{S},P)+(1-w){\epsilon}_\theta({z}_t,t),
\end{equation}
where $w$ represents the classifier-free guidance factor (CFG).

%#########################################################%

\section{Experiments}
\subsection{Experimental Setup}
\myparagraph{Setup.} For the IMAGstyle dataset,  during the training phase,  we suggest using `a [vcp]' as a prompt for content images and `a [stp]' as a prompt for style images. The rank is set to 64 and each B-loRA is trained with 1000 steps.
 % Once trained, the combination of separated content LoRA and style LoRA yields styled results.
 %
 During the generation phase, we suggest using `a [vcp] in [stv] style' as the prompt.
 For the CSGO framework, we employ \textit{stabilityai/stable-diffusion-xl-base-1.0} as the base model, pre-trained \textit{ViT-H} as image encoder, and \textit{TTPlanet/TTPLanet\_SDXL\_Controlnet\_Tile\_Realistic} as ControlNet. we uniformly set the images to $512 \times 512$ resolution. The drop rate of text, content image, and style image is 0.15. The learning rate is 1e-4.  
During training stage, $\lambda_c =\lambda_s = \delta_c=1.0$.  During inference stage, we suggest $\lambda_c =\lambda_s=1.0$ and $\delta_c=0.5$.
Our experiments are conducted
on 8 NVIDIA H800 GPUs (80GB) with a batch size of 20 per GPU and trained 80000 steps. 

\myparagraph{Datasets and Evaluation.} We use the proposed IMAGStyle as a training dataset and use its testing dataset as an evaluation dataset. %
We use the CSD score~\cite{somepalli2024measuringstylesimilaritydiffusion} as an evaluation metric to evaluate the style similarity.
Meanwhile, we employ the proposed content alignment score (CAS) as an evaluation metric to evaluate the content similarity.

% \myparagraph{Quantitative results.}
\begin{table}[t]
\scriptsize
  \centering
  \caption{Comparison of style similarity (CSD) and content alignment (CAS) with recent state-of-the-art methods on the test dataset.}\renewcommand{\arraystretch}{1.15}
  \begin{adjustbox}{width=1\textwidth}
  \begin{tabular}{c|ccccccc} 
    \toprule[1.5pt]
    & StyTR$^2$ & Style-Aligned & StyleID & InstantStyle & StyleShot & StyleShot-lineart & CSGO \\
    &\cite{deng2022stytr2}&\cite{hertz2024style}& \cite{chung2024style} & \cite{wang2024instantstyle} & \cite{gao2024styleshot} & \cite{gao2024styleshot} & Ours\\
    \midrule % 中间横线
    CSD ($\uparrow$) & 0.2695 &0.4274&   0.0992 &0.3175&0.4522&   0.3903 &0.5146\\
    
    CAS ($\downarrow$) & 0.9699 &1.3930  &0.4873 &1.3147 &  1.5105&   1.0750 &0.8386\\
 
    \bottomrule % 下边框
  \end{tabular}
  \end{adjustbox}
  \label{tab1}
  \vspace{-0.3cm}
\end{table}

\begin{figure}[h!]
    \centering\setlength{\abovecaptionskip}{2pt}
    \includegraphics[width=.99\linewidth]{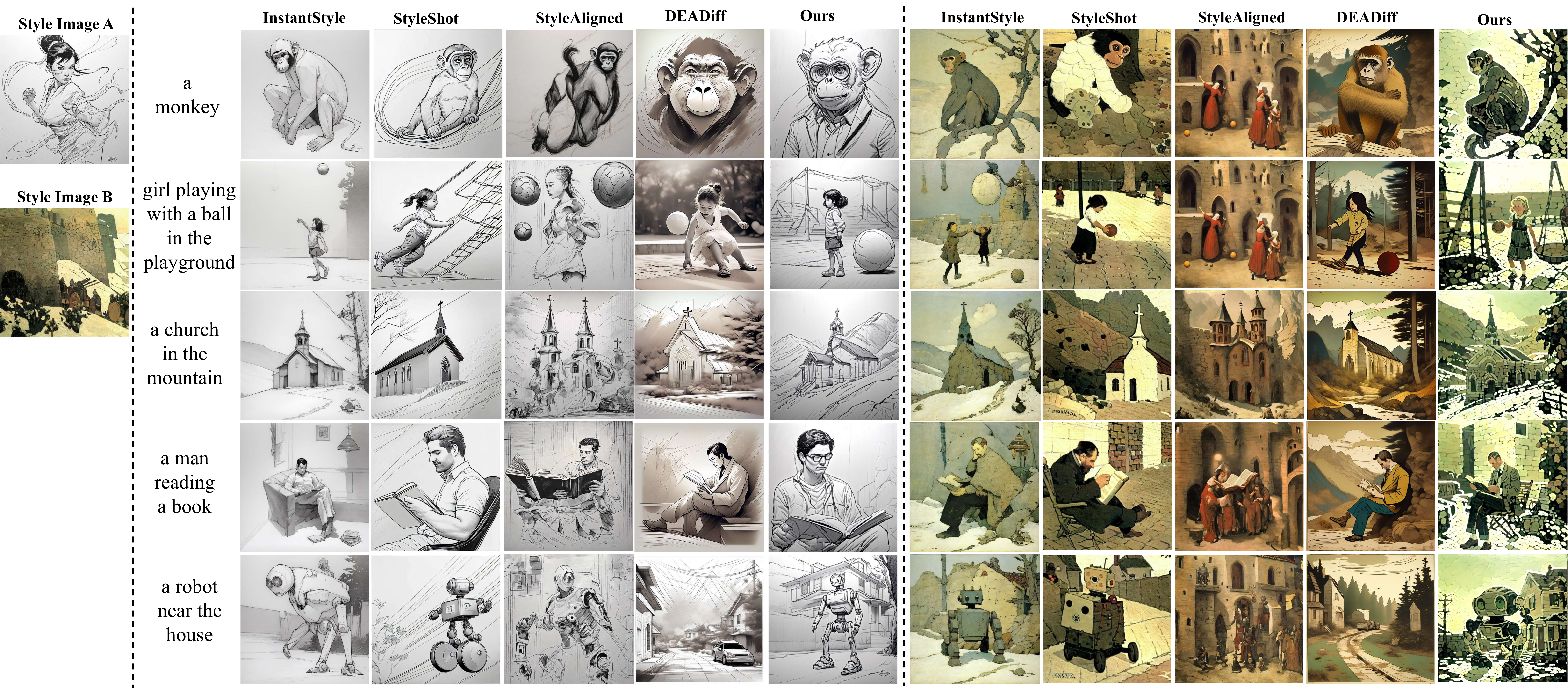}
    \caption{Comparison of generation results for text-driven stylized synthesis with recent methods.}
    \label{fig:4}
    \vspace{-0.3cm}
\end{figure}
% \begin{figure}
%     \centering
%     \includegraphics[width=1\linewidth]{pictures/new_6.pdf}
%     \caption{Enter Caption}
%     \label{fig:enter-label}
% \end{figure}

\begin{figure}[h!]
% \centering
\begin{minipage}[t]{0.43\textwidth}

\includegraphics[width=.97\linewidth]{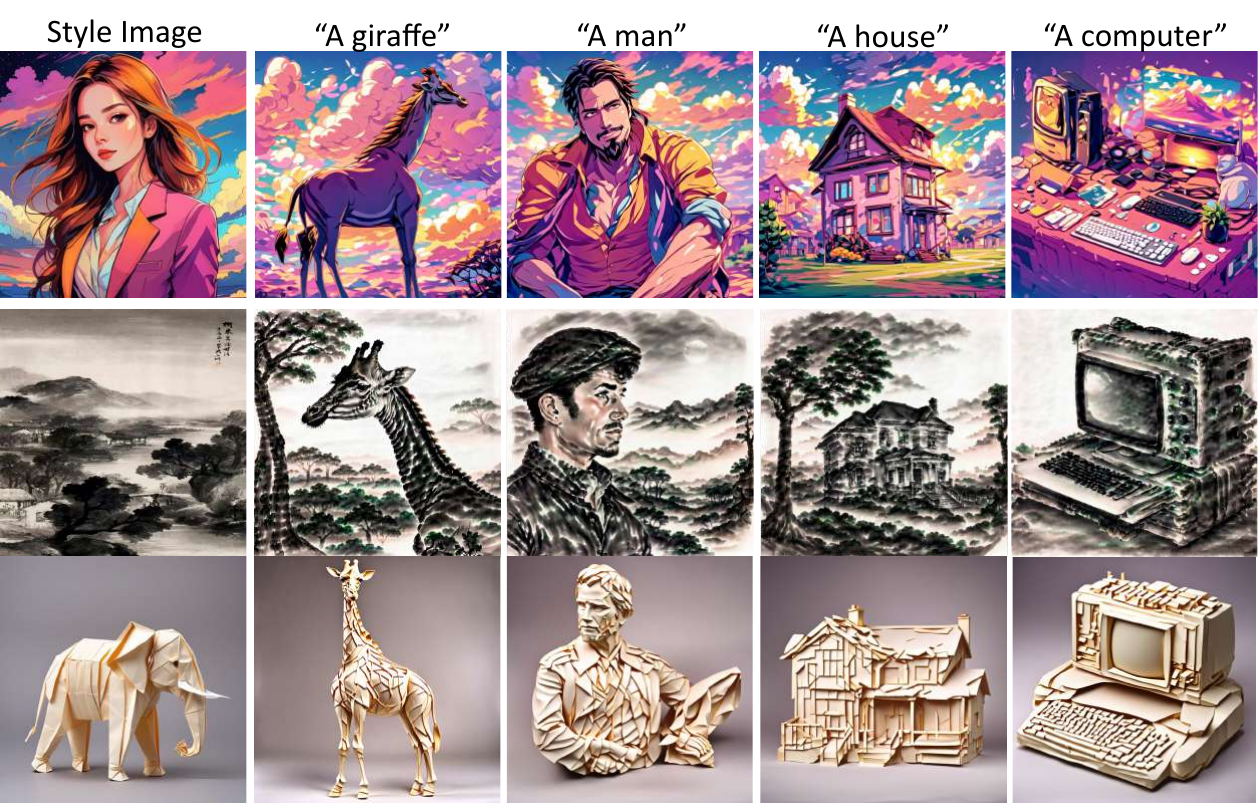}
\centering\setlength{\abovecaptionskip}{1pt}
	\caption{%
    Generated results of the proposed CSGO in text-driven stylized synthesis. 
    }
	\label{fig5}
\end{minipage}
\begin{minipage}[t]{0.01\textwidth}
\end{minipage}
\begin{minipage}[t]{0.56\textwidth}
\centering
 \includegraphics[width=.98\linewidth]{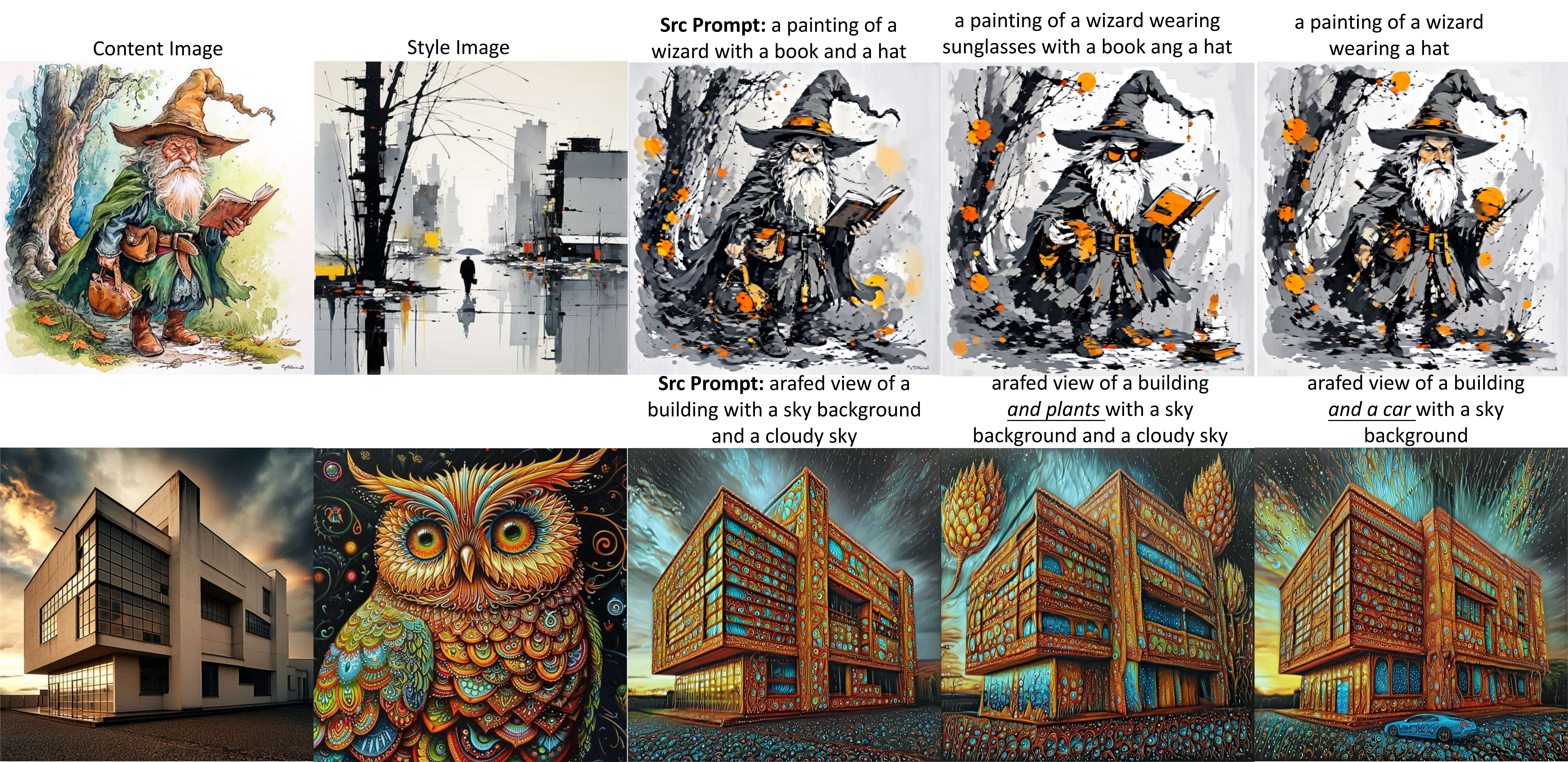}\setlength{\abovecaptionskip}{1pt}
    \caption{The generated results of the proposed CSGO in text editing-driven stylized synthesis. }
    \label{fig6}
\end{minipage}
\vspace{-0.2cm}
\end{figure}

\myparagraph{Baselines.}
% \myparagraph{Baselines.} 
We compare recent advanced inversion-based StyleID~\cite{chung2024style}, StyleAligned~\cite{hertz2024style} methods, and StyTR$^2$~\cite{deng2022stytr2} based on the Transformer structure. In addition, we compare Instantstyle~\cite{wang2024instantstyle} and StyleShot (and their fine-grained control method StyleShot-lineart)~\cite{gao2024styleshot} that introduce ControlNet and IPAdapter structures as baselines. For text-driven style control task, we also introduce DEADiff~\cite{qi2024deadiff} as a baseline.
% \begin{figure}
%     \centering
%     \includegraphics[width=1\linewidth]{pictures/new_8.pdf}
%     \caption{Enter Caption}
%     \label{fig:enter-label}
% \end{figure}
\subsection{Experimental Results}
\myparagraph{Image-Driven Style Transfer.} 
In Table~\ref{tab1}, we demonstrate the CSD scores and CAS of the proposed method with recent advanced methods for the image-driven style transfer task.
In terms of style control, our CSGO achieves the highest CSD score, demonstrating that CSGO achieves state-of-the-art style control.
Due to the decoupled style injection approach, the proposed CSGO effectively extracts style features and fuses them with high-quality content features.
As illustrated in Figure~\ref{fig:3}, Our CSGO precisely transfers styles while maintaining the semantics of the content in natural, sketch, face, and art scenes.

% In Figure~\ref{fig:3}, we show the style transfer results of the proposed CSGO in different scenes and compare them with recent state-of-the-art approaches.

\begin{figure}[t]
% \centering
\begin{minipage}[t]{0.6\textwidth}
\centering\setlength{\abovecaptionskip}{2pt}
\includegraphics[width=.99\textwidth]{pictures_process/new_9.pdf}
	\caption{%
    Ablation studies of content control and style control.
    }
	\label{fig:7}
\end{minipage}
\begin{minipage}[t]{0.1\textwidth}
\end{minipage}
\begin{minipage}[t]{0.39\textwidth}
\centering
\setlength{\abovecaptionskip}{1.5pt}
 \includegraphics[width=.98\textwidth]{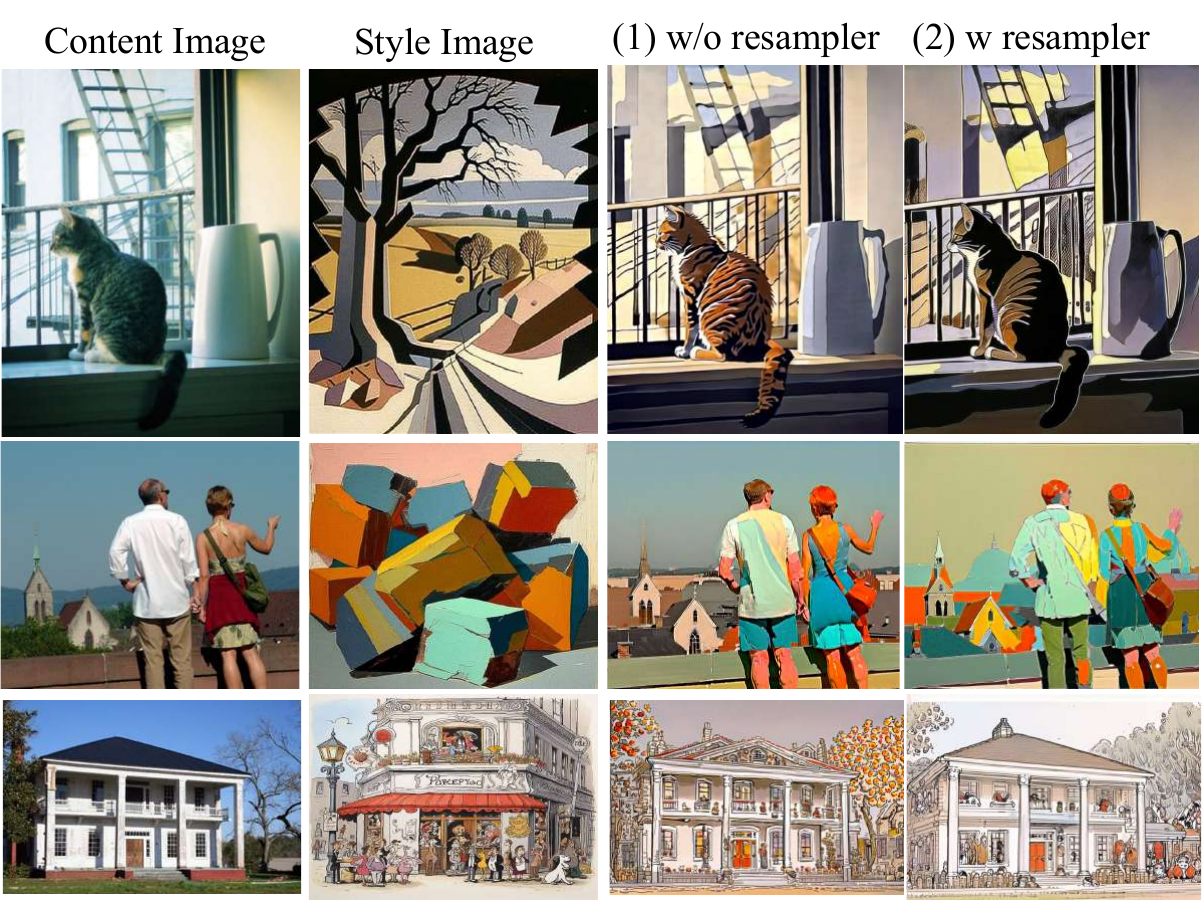}
 % \vspace{-0.2cm}
    \caption{Ablation studies of style image projection. 
    }
    \label{fig8-1}
\end{minipage}
\vspace{-0.4cm}
\end{figure}

\begin{figure}[h!]
    % \vspace{-0.3cm}
    \centering\setlength{\abovecaptionskip}{0pt}
    \includegraphics[width=.99\linewidth]{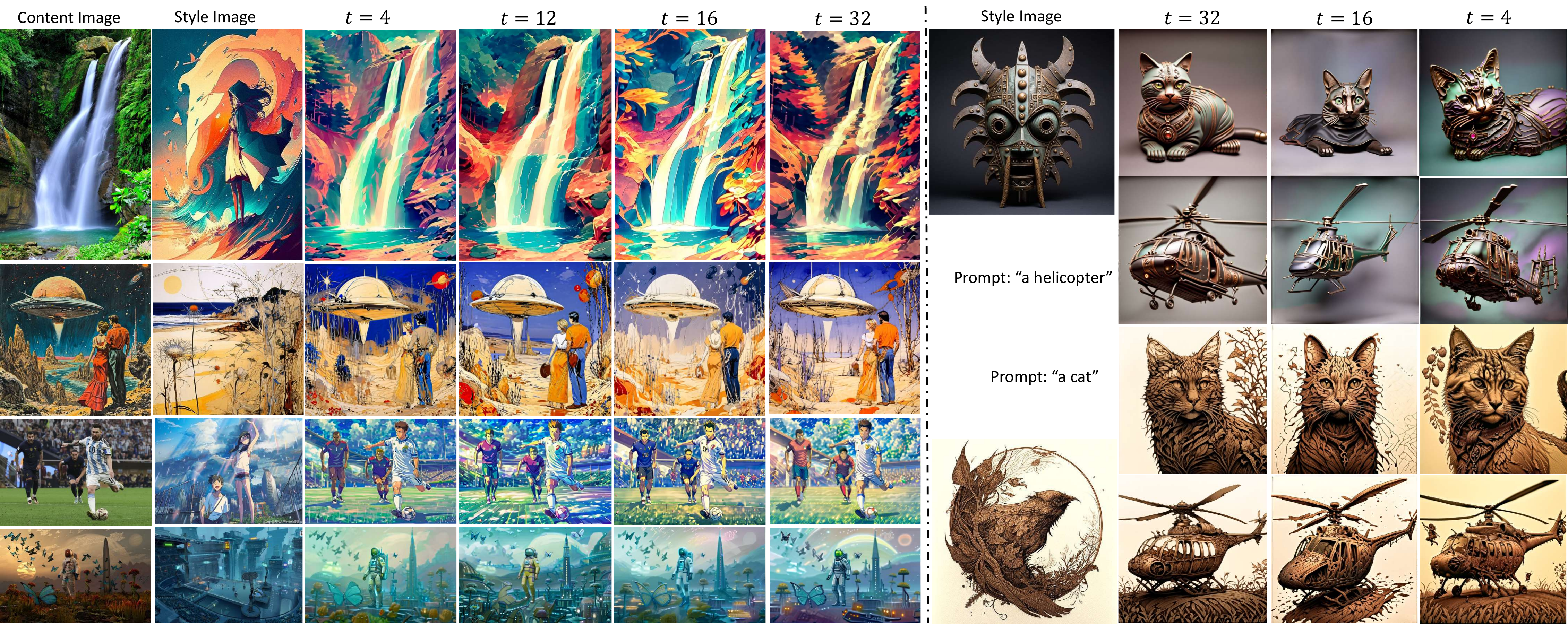}
    \caption{Ablation studies of style token number $t$. Left: Image style transfer results. Right: Text-driven stylized synthesis results.}
    \label{fig11}
    \vspace{-0.4cm}
    % \end{minipage}
\end{figure}

In terms of content retention, it can be observed that StyleID~\cite{chung2024style} and StyleAligned~\cite{hertz2024style}, which are based on inversion, maintain the original content too strongly in sketch style transfer scenarios (CAS is very low). However. they are unable to inject style information since CSD score is low. InstantStyle~\cite{wang2024instantstyle} and StyleShot~\cite{gao2024styleshot} (including Lineart), which use lines to control the content, are affected by the level of detail of the lines and have different degrees of loss of content (such as face scenes).
The proposed CSGO directly utilizes all the information of the content image, and content preservation is optimal.
The quantitative results in Table~\ref{tab1} also show that the proposed CSGO maintains high-quality content retention with precise style transfer.
%

% \begin{figure}
%     % \vspace{-0.cm}
%     \centering\setlength{\abovecaptionskip}{2pt}
%     \includegraphics[width=.6\linewidth]{pictures/new_9.pdf}
%     \caption{Ablation studies of Controlnet injection and IP-Adapter injection.}
%     \label{fig:7}
%     \vspace{-0.2cm}
% \end{figure}

%

\myparagraph{Text-Driven Stylized Synthesis.} The proposed method enables text-driven style control, \textit{i.e.}, given a text prompt and a style image, generating images with similar styles.
Figure~\ref{fig:4} shows the comparison of the generation results of the proposed CSGO with the state-of-the-art methods.
In a simple scene, it is intuitive to observe that our CSGO obeys textual prompts more. The reason for this is that thanks to the explicit decoupling of content and style features, style images only inject style information without exposing content.
In addition, in complex scenes, thanks to the well-designed style feature injection block, CSGO enables optimal style control while converting the meaning of text. As illustrated in Figure~\ref{fig5}, we demonstrated more results.
%
% \begin{wrapfigure}{r}{0.55\textwidth}
%     \vspace{-0.3cm}
%     \centering\setlength{\abovecaptionskip}{0pt}
%     \includegraphics[width=.99\linewidth]{pictures/new_9.pdf}
%     \caption{Ablation studies of Controlnet injection and IP-Adapter injection.}
%     \label{fig:7}
%     \vspace{-0.5cm}
%     % \end{minipage}
% \end{wrapfigure}

\myparagraph{Text editing-Driven Stylized Synthesis.} The proposed CSGO supports text editing-driven style control. As shown in Figure~\ref{fig6}, in the style transfer, we maintain the semantics and layout of the original content images while allowing simple editing of the textual prompts. The above excellent results demonstrate that the proposed CSGO is a powerful framework for style control.

\vspace{-0.25cm}
\subsection{Ablation Studies.}
\begin{wrapfigure}{r}{0.6\textwidth}
\vspace{-0.45cm}
    \centering\setlength{\abovecaptionskip}{0pt}
    \includegraphics[width=0.95\linewidth]{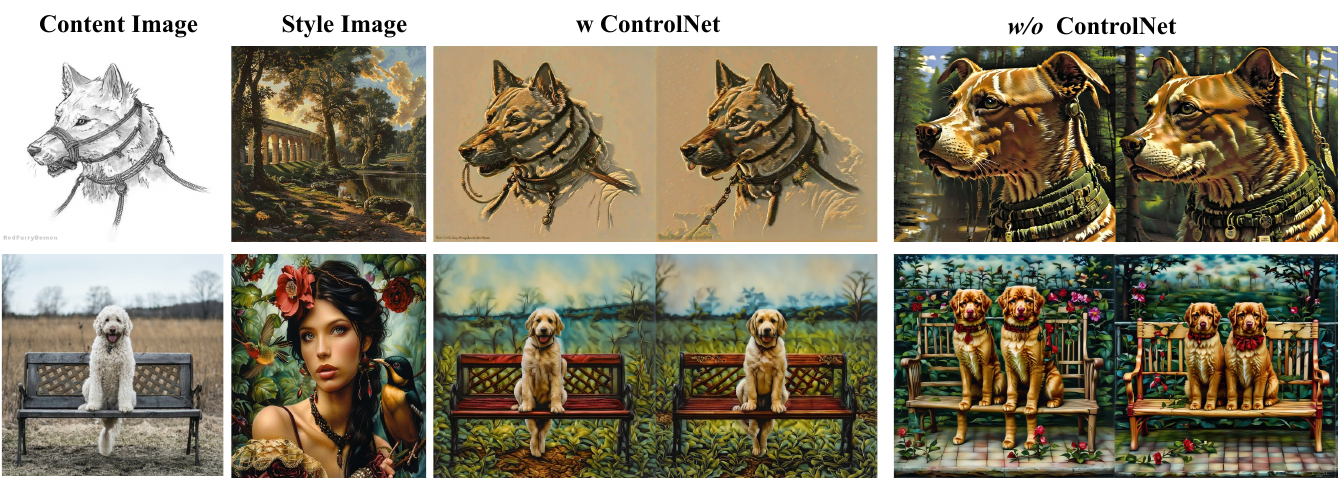}
    \caption{Ablation studies of ControlNet. }
    \vspace{-0.35cm}
    \label{fig:12}
\end{wrapfigure}

\myparagraph{Content control and style control.} 
We discuss the impact of the two feature
% \begin{wrapfigure}{r}{0.5\textwidth}
% \vspace{-0.45cm}
%     \centering\setlength{\abovecaptionskip}{0pt}
%     \includegraphics[width=0.95\linewidth]{pictures/new_14.pdf}
%     \caption{Ablation studies of ControlNet. }
%     \vspace{-0.35cm}
%     \label{fig:12}
% \end{wrapfigure}
%
 injection methods, as shown in Figure~\ref{fig:7}.
The content image must be injected via ControlNet injection to maintain the layout while preserving the semantic information. 
If content images are injected into the base model only through an additional Cross attention layer, only semantic information is guaranteed, while the full content information is not preserved (Figure~\ref{fig:7}(1)). 
After introducing the ControlNet injection, the quality of content retention improved, as shown in Figure~\ref{fig:12}. However, if the style features are injected into base UNet only without ControlNet injection, this weakens the style of the generated images, which can be observed in the comparison of Figure~\ref{fig:7}(2) and (3). Therefore, the proposed CSGO pre-injects style features in the ControlNet branch to further fuse the style features to enhance the transfer effect.

\myparagraph{Style image projection layer.} The style image projection layer can effectively extract style features from the original embedding. We explore the normal linear layer and the Resampler structure, and the experimental results are shown in Figure~\ref{fig8-1}. Using the Resampler structure captures more detailed style features while avoiding content leakage.

\myparagraph{Token number.} We explore the effect of the number of token $t$ in the style projection layer on the results of style transfer and text-driven style synthesis. The experimental results are shown in Figure~\ref{fig8-1}, where the style control becomes progressively better as $t$ increases. This is in line with our expectation that $t$ influences the quality of feature extraction.  A larger $t$ means that the projection layer can extract richer style features. 
%
% In our experiment, $t$ did not converge with the scale-up feature.

% \newpage

% \begin{figure}[t]
% % \centering
% \begin{minipage}[t]{0.67\textwidth}
% \centering\setlength{\abovecaptionskip}{2pt}
% \includegraphics[width=.98\textwidth]{pictures/new_11.pdf}
% 	\caption{%
%     Ablation studies of style token number $t$.
%     }
% 	\label{fig11}
% \end{minipage}
% \begin{minipage}[t]{0.1\textwidth}
% \end{minipage}
% \begin{minipage}[t]{0.32\textwidth}
% \centering
% \setlength{\abovecaptionskip}{1.5pt}
%  \includegraphics[width=.98\textwidth]{pictures/new_7.pdf}
%  % \vspace{-0.2cm}
%     \caption{Ablation studies of content scale $\delta_c$, CFG, and style scale $\lambda_S$.
%     }
%     \label{fig:8}
% \end{minipage}
% \vspace{-0.4cm}
% \end{figure}

% \begin{wrapfigure}{r}{0.5\textwidth}
% % \begin{figure}
% % \begin{minipage}
% \vspace{-0.2cm}
%     \centering\setlength{\abovecaptionskip}{0pt}
%     \includegraphics[width=.49\textwidth]{pictures/new_7.pdf}
%     % \vspace{-6pt}
%     \caption{Ablation studies of content scale $\delta_c$, CFG, and style scale $\lambda_S$.}
%     \label{fig:8}
%     \vspace{-1.6cm}
%     % \end{minipage}
% \end{wrapfigure}

\begin{figure}[h]
    % \vspace{-0.3cm}
      \centering\setlength{\abovecaptionskip}{2pt}
    \includegraphics[width=.99\textwidth]{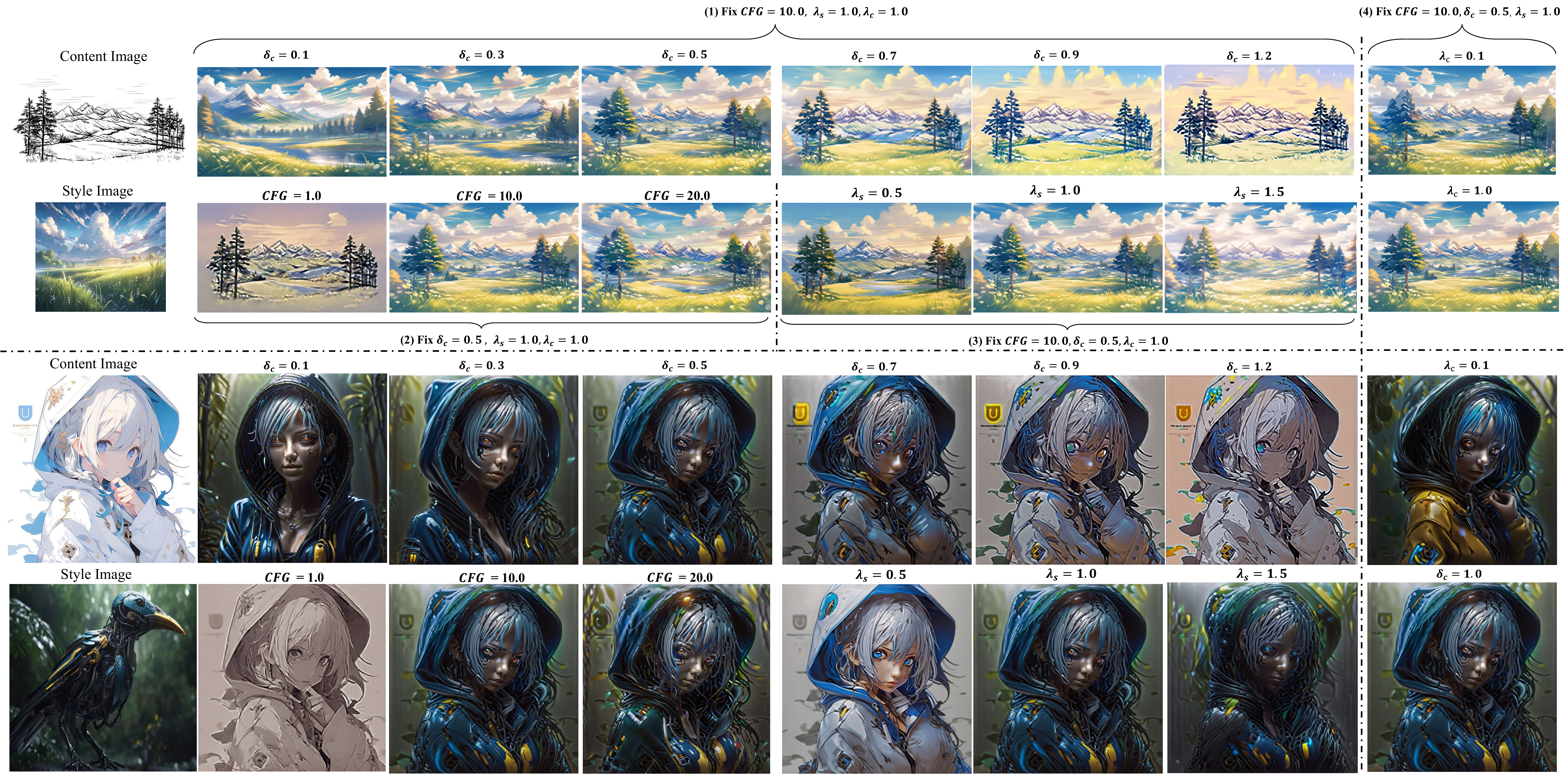}
    % \vspace{-6pt}
    \caption{Ablation studies of content scale $\delta_c$, CFG, content scale $\lambda_c$, and style scale $\lambda_s$.}
    \label{fig:8}
    \vspace{-0.3cm}
    % \end{minipage}
\end{figure}
\myparagraph{The impact of content scale $\delta_c$.} As shown in Figure~\ref{fig:8}, when $\delta_c$ is small, the content feature injection is weak, and CSGO obeys the textual prompts and style more. As $\delta_c$ increases, the quality of content retention becomes superior. However, we notice that when $\delta_c$ is large (e.g., 0.9 and 1.2), the style information is severely weakened.

\myparagraph{The impact of CFG scale.} Classifier-free guidance enhances the capabilities of the text-to-image model. The proposed CSGO is similarly affected by the strength of CFG scale. As shown in Figure~\ref{fig:8}, the introduction of CFG enhances the style transfer effect.

\myparagraph{The impact of style scale $\lambda_s$ and content scale $\lambda_c$.} The style scale affects the degree of style injection. Figure~\ref{fig:8} shows that if the style scale is less than 1.0, the style of the generated image is severely weakened. We suggest that the style scale should be between 1.0 and 1.5. Content control in the down-sampling blocks utilizes the semantic information of the content image to reinforce the accurate retention of content. Figure~\ref{fig:8} shows that $\lambda_c$ is most effective when it is near 1.0.

% \myparagraph{dropout}

% \myparagraph{conditional scale}

\section{Conclusion.}

We first propose a pipeline for the construction of content-style-stylized image triplets. Based on this pipeline, we construct the first large-scale style transfer dataset, IMAGStyle, which contains 210K image triplets and covers a wide range of style scenarios. To validate the impact of IMAGStyle on style transfer, we propose CSGO, a simple but highly effective end-to-end training style transfer framework, and we verify that the proposed CSGO can simultaneously perform image style transfer, text-driven style synthesis, and text editing-driven style synthesis tasks in a unified framework. Extensive experiments validate the beneficial effects of IMAGStyle and CSGO for style transfer. 
We hope that our work will inspire the research community
to further explore stylized research.

\myparagraph{Future work.} Although the proposed dataset and framework achieve very advanced performance, there is still room for improvement. Due to time and computational resource constraints, we constructed only 210K data triplets. We believe that by expanding the size of the dataset, the style transfer quality of CSGO will be even better. Meanwhile, the proposed CSGO framework is a basic version, which only verifies the beneficial effects of generative stylized datasets on style transfer. We believe that the quality of style transfer can be further improved by optimizing the style and content feature extraction and fusion methods.

% \subsection{Limitation.}
% \subsection{Social Influence.}

\bibliography{iclr2024_conference}
\bibliographystyle{iclr2024_conference}

\appendix
% \section{Appendix}

\end{document}

\maketitle
In this supplementary material, we show more ablation experiments and more style control results.

\myparagraph{Ablation Study on ControlNet.}

\bibliography{iclr2024_conference}
\bibliographystyle{iclr2024_conference}

\appendix
% \section{Appendix}